\documentclass[conference, 10pt]{IEEEtran}
\IEEEoverridecommandlockouts

\usepackage{amsmath}
\usepackage{graphicx, graphics, theorem, times, amsfonts, amsmath, amssymb, cite}
\usepackage{tikz}
\usetikzlibrary{shapes,arrows}
\usepackage{pgfplots}
\usepackage{color}
\usepackage{hyperref}
\usepackage{multirow}
\usepackage{rotating}
\usepackage{cleveref, subcaption}
\usepackage{bm}
\input{mysymbol.sty}
\usepackage[utf8]{inputenc}
\usepackage{graphicx}

\def\Tr{\mathsf{T}}

\newcommand{\QED}{\hfill\ensuremath{\blacksquare}}

\title{Low-rank$\hspace{.1cm}$State-action$\hspace{.1cm}$Value-function$\hspace{.1cm}$Approximation}

\author{Sergio Rozada, Victor Tenorio,  and Antonio G. Marques$^{*}$ \vspace{-.13cm}
	\thanks{This work was supported by the Spanish Federal Grants SPGRAPH (PID2019-105032GB) and by the Young Researchers R\&D Project Grants F661-MAPPING-UCI and F663-AAGNCS both funded by the Community of Madrid (CAM) and the King Juan Carlos University (URJC). All the authors are with the Dept. of Signal Theory and Comms., URJC, Madrid, Spain. Sergio Rozada is also with Clients Solutions Advanced Analytics, BBVA, Madrid, Spain. $^{*}$Contact author: antonio.garcia.marques (AT) urjc.es.}
    }

\vspace{-.2cm}
\begin{document}
	%\ninept

\maketitle
\vspace{-.2cm}

%%%%%%%%%%%%%%%%
\begin{abstract}
	Value functions are central to Dynamic Programming and Reinforcement Learning but their exact estimation suffers from the curse of dimensionality, challenging the development of practical value-function (VF) estimation algorithms. Several approaches have been proposed to overcome this issue, from non-parametric schemes that aggregate states or actions to parametric approximations of state and action VFs via, e.g., linear estimators or deep neural networks. Relevantly, several high-dimensional state problems can be well-approximated by an intrinsic low-rank structure. Motivated by this and leveraging results from low-rank optimization, this paper proposes different stochastic algorithms to estimate a low-rank factorization of the $Q(s, a)$ matrix. This is a non-parametric alternative to VF approximation that dramatically reduces the computational and sample complexities relative to classical $Q$-learning methods that estimate $Q(s,a)$ separately for each state-action pair. 
	%Numerical results in two standard RL test-cases demonstrate that our algorithms converge faster and achieve a higher reward than other non-parametric alternatives. 
\end{abstract}

\section{Introduction and motivation}
\vspace{-.1cm}
As complex interconnected systems and big data become pervasive, the engineering goal is to design intelligent algorithms that leverage the data and learn how to interact with the world autonomously. Reinforcement Learning (RL) tries to embed into algorithms how humans interact with the world, learning in real-time by trial and error \cite{sutton2011reinforcement,bertsekas2019reinforcement,mnih2013playing}. Technically speaking, RL solves sequential optimization problems, like Dynamic Programming (DP) \cite{bertsekas2019reinforcement}, in a model-free fashion. When the environment is ``simple'', solid theoretical results exist, and a range of algorithms address how to learn the mapping from the state observed at every time instant to the action to be executed. However, RL is algorithmically challenging when the actions and states describing the environment are numerous, high-dimensional, or defined in continuous domains. Such a curse of dimensionality calls for approximated algorithms that, e.g., discretize continuous domains while simultaneously limiting the complexity of the problem to be solved \cite{sutton2011reinforcement}. Indeed, different (parametric and nonparametric) approaches to reduce the number of degrees of freedom (keeping computational complexity under control and facilitating learning) at the cost of sacrificing optimality exist \cite{bertsekas2019reinforcement}.  
Motivated by its practical relevance and the described computational challenges, this paper puts forth  nonparametric, stochastic, model-free algorithms that leverage results from low-rank optimization and matrix completion \cite{eckart1936approximation,markovsky2012low}. The rest of this section provides details on the notation and fundamentals of RL and, once those have been presented, describes the contribution in a more rigorous manner. 
%Sec.2 introduces the different algorithms proposed, all designed as stochastic schemes associated with different low-rank optimizations. Sec.3 presents numerical results testing some of the proposed algorithms in a couple of classical RL benchmark scenarios. Short concluding remarks wrap up this paper in Sec.4.   

\vspace{.05cm}
\noindent \textbf{Fundamentals of RL and notation.}
We deal with closed-loop setups where agents interact sequentially with an environment. The three key elements to define the RL problem are the states describing the environment, the actions agent(s) can take, and their associated rewards. Mathematically, the environment is represented by a (discretized) set of states $\ccalS$. In each state, the agent takes actions from a (discretized) set of actions, or action space, $\ccalA$. 
%Although many problems deal with discrete state and action spaces, more challenging continuous formulations are of interest too. 
We assume a time-slotted scenario and, with $t=1,...,T$ representing the time index.
After taking an action $a_t$ in a particular state $s_t$, the agent receives a numerical signal, or reward $r_t$, that quantifies the instantaneous value of that state-action pair. Two key aspects in RL are: (i) the dependence of $r_t$ on $s_t$ and $a_t$ is oftentimes not deterministic, and (ii) the action $a_t$ has an impact not only in $r_t$ but also in $s_{t'}$ for $t'>t$, coupling the optimization across time. In other words, when deciding the value of $a_t$ one must take into account not only its impact on $r_t$ but also on $r_{t'}$ for $t>t'$. A common approach to deal with these two issues is to rely on Markovianity, recast first the optimization within the framework of Markov Decision Process (MDP) and, then, view RL as an stochastic approach for the MDP. 
%To that end, we consider the expected reward value and the probability state-transition function 
%\vspace{-.8cm} 
%
%\begin{align}
%\ccalR^a_{s s^{'}} &= \mathbb{E}[r_{t+1}=r|s_t\!=\!s, a_t\!=\!a, s_{t+1}\!=\!s']\;\;\text{and}\;\;\nonumber\\
%\ccalP^a_{s s^{'}} &= \Pr{s_{t+1}=s'|s_t\!=\!s, a_t\!=\!a},\label{eq:definitions_MDP}  %\vspace{-.8cm} 
%\end{align}
%
%where $\ccalP^a_{s s^{'}}$ models the dynamics of the system once a particular state-action pair is executed and $s'$ denotes the state at the next time instant. These four elements form the Markov Decision Process (MDP) tuple $(\ccalS, \ccalA, \ccalP^a_{s s^{'}}, \ccalR^a_{s s^{'}})$, which is a convenient framework to formalize the RL problem.

In this context, suppose that we have a policy $\pi:\ccalS\mapsto\ccalA$ that maps states into actions and define the expected aggregated reward as $\mathbb{E}[\sum_{\tau=t}^{T}\gamma^{\tau-t}r_\tau|s_t,a_t]$, where $\gamma\in(0,1)$ is a discount factor that places more focus on near-future values, and the time-horizon is oftentimes set to  $T=\infty$ \cite{sutton2011reinforcement,bertsekas2019reinforcement}. Given the policy $\pi$, the so-called value function (VF) quantifies the expected reward associated with that particular (alternatively, the optimal) policy. Since the expectation depends on both the state and the action, two versions of the VF can be defined. For the one used in this paper, the so-called state-action value function (SA-VF) $Q^\pi:\;\ccalS\times \ccalA\rightarrow \reals$, the dependence on the state and the action is kept. 
%In the second one, we use the fact that the action is a function of the state to define the state VF $V^\pi:\;\ccalS \rightarrow \reals$. 
Since each $(s,a)$ pair has one $Q$-value associated with it, when the states and actions are discrete, all the $Q$-values are arranged in the form of a matrix $\bbQ^\pi\in\reals^{D_\ccalS\times D_\ccalA}$, with $D_S$ being the number of states (the cardinality of $\ccalS)$ and $D_A$ the cardinality of $\ccalA$ \cite{sutton2011reinforcement}. When the state (action) is $d_S$-dimensional and continuous, the classical approach is to discretize each dimension using $N_S$ intervals so that $D_S=N_S^{d_S}$. This exponential dependence affects computational and learning performance, calling for algorithms to keep the complexity of the model under control.   
%Similarly, one can define the vector $\bbV^\pi\in \reals^{D_\ccalS}$. 
%
%Generally speaking, there are two main families of approaches to learning $\pi$, those where the goal is to learn (estimate) $\pi$ directly and those where, leveraging the Bellman equations \cite{bellman1957dynamic}, $\pi$ is defined based on the VF and, as a result, the goal is to design iterative algorithms to estimate the VF itself \cite{sutton2011reinforcement}. 
%Policy-iteration methods belong to the first family, while value-iteration methods belong to the second one \cite{sutton2011reinforcement}. 
%
%The estimation can be handled using parametric or nonparametric approaches, with a broad range of algorithms existing for any of them. 
%If, for example, the approach is to estimate the policy under a parametric approach, then $\pi$ is replaced by $\pi_{\theta}$, with vector $\theta$ containing the parameters describing the policy, and the focus shifts to estimating $\bbtheta$. 

%The goal in RL is also to learn a policy $\pi:\ccalS\mapsto\ccalA$ that maximizes the expected aggregated reward. The challenges include the facts that (i) the objective in RL is to learn a mapping and, hence, the number of parameters to estimate grows with both $D_{\ccalS}$ and $D_{\ccalA}$ and (ii).
The distinctive feature in RL is that the statistical dependence across states and actions is unknown (or intractable), and one must resort to model-free algorithms that learn on-the-fly by using realizations (trajectories) of the MDP \cite{sutton2011reinforcement,bertsekas2019reinforcement}. As in MDP and DP, there are RL methods focused on learning the policy $\pi$ --typically under a parametric approach followed by stochastic gradient descent (SGD)--, while others focus on learning the VF (using parametric or nonparametric approaches). Stochastic optimization of neural networks (NN) and linear-based models are widely-used parametric alternatives \cite{sutton2011reinforcement,bertsekas2019reinforcement}, while the classical $Q$-learning algorithm is the most celebrated example of the non-parametric class \cite{watkins1992q}.

\vspace{.05cm}
\noindent \textbf{Contribution and related work.} This manuscript proposes different \emph{low-rank stochastic} RL algorithms to estimate the $Q$-function in a \emph{model-free} and \emph{online} setup. Most of the proposed algorithms are designed using SGD and can be interpreted as temporal differences (TD)-based schemes \cite{bertsekas2019reinforcement}. Once the estimation concludes, the policies are obtained as those that, for the current state, select the action maximizing the estimated SA-VF. The key aspect of the design is to regularize the estimation problem by promoting an SA-VF with a low-rank structure via matrix factorization. 
%Two techniques are used to that end: matrix factorization, and matrix factorization with Frobenius regularizers. 
Those techniques have been successfully utilized in the context of low-rank optimization and matrix completion \cite{eckart1936approximation,markovsky2012low,udell2016generalized,mardani2013decentralized}, but their use in DP/MDP (and, especially, in RL and TD) has been limited. For example, \cite{barreto2016incremental,jiang2017contextual} used matrix-factorization approaches to approximate the transition matrix of an MDP and, then, leverage those to obtain the associated VF and optimal policy. In \cite{ong2015value}, the author proposes obtaining first the $Q$-function of an MDP and, then, approximate it using a low-rank plus sparse decomposition. Similarly, in the context of DPs associated with energy storage, \cite{cheng2016co, cheng2017low} proposed a low-rank (rank-one) approximation for the estimation of the state VF. Recently, low-rank optimization has been proposed to approximate the SA-VF using \emph{model-based} and \emph{offline} setups \cite{yang2019harnessing, shah2020sample}. Another work that uses factorization techniques in the context of RL is \cite{chen2018factorized}, where the focus is on deep factorized architectures amenable to be implemented distributedly in multi-agent setups. Also within RL setups, \emph{linear models} have been used to approximate the SA-VF on-the-fly \cite{melo2007q}. In those works a set of features is defined (based on prior knowledge, collected data, or spectral properties of the transition matrices \cite{behzadian2019fast, behzadian2018low}) and then each entry of the SA-VF is modeled as a weighted sum of the features associated with the $(s,a)$ pair. The weights are assumed to be the same across $Q$ and the focus of the RL algorithm is on their estimation using sample trajectories. As discussed in detail at the end of Sec. 2, a distinctive feature of the low-rank RL approaches we put forth in this paper is that, here, the features are assumed to follow a factorized (bilinear) model and both features and weights are learned jointly, in real-time, from the observed data.

\section{Approximate low-rank state-action value function estimation}
\vspace{-.1cm}

Before presenting our algorithms, we need to provide a succinct description of classical nonparametric RL estimation of the $Q$-function, %, discusses two alternatives to promote a low-rank structure in the estimated SA-VF, and presents different stochastic optimization schemes to handle the associated estimation problems. 
which tries to optimize:
\begin{eqnarray}\label{E:Q_estimation_plain}
&\hbQ = \argmin_{\bbQ} \frac{1}{2}\sum_{(s,a)\in \ccalM} (q_s^a - [\bbQ]_{s,a})^2&
\end{eqnarray}
where $\ccalM$ is the set of state-action tuples $(s, a)$ that the agent has sampled from the environment, and $q_s^a$ is the target signal when being in state $s$ and taking action $a$. In contrast to other learning paradigms, in RL this target signal $q_s^a$ is not known in advance and should be estimated on the fly. To estimate the target signal, we can use the available reward to define $q_s^a=r_s^a+\gamma [\hbQ]_{s',a'}$, where the tuple $(s',a')$ represents the state-action pair sampled in the subsequent time instant. 

Suppose now that: i) the action $a'$ is selected as the one maximizing the current estimate of the SA-VF and, hence, $[\hbQ]_{s',a'}=\max_a [\hbQ]_{s',a}$; and ii) \eqref{E:Q_estimation_plain} is solved via an SGD method that, at each time $t$, updates the current estimates based on $(s_t,a_t,r_t,s_{t+1})$. This yields the following update rule
\begin{align}
[\hbQ^t]_{s_t,a_t} =&  [\hbQ^{t-1}]_{s_t,a_t} + \alpha_t \big( r_t + \gamma \max_a [\hbQ^{t-1}]_{s_{t+1},a}   \nonumber\\
 -& [\hbQ^{t-1}]_{s_t,a_t}\big)\label{E:Q_learning_estimation}
\end{align}
with $\alpha_t$ being the learning rate (stepsize), $\hbQ^{t}$ being the estimate of the SA-VF at time $t$, and where we remark that $[\hbQ^t]_{s,a}=[\hbQ^{t-1}]_{s,a}$ for all $(s,a)\neq (s_t,a_t)$. The scheme in \eqref{E:Q_learning_estimation} corresponds to the celebrated (TD-based) $Q$-learning algorithm, which enjoys convergence guarantees, provided that all $(s,a)$ pairs are visited infinitely often \cite{watkins1992q}. To facilitate this latter point, an exploration module is added to the algorithm so that with probability $(1-\varepsilon_t)$ the action $a'=\argmax_{a} [\hbQ^{t-1}]_{s_{t+1},a} $  is selected (and, hence, the expression in \eqref{E:Q_learning_estimation} applies) and, with probability $\varepsilon_t$, the action $a'$ is chosen uniformly at random. 
%The exploration probability $\varepsilon_t$ is typically set to a medium-small value that decreases as $t$ increases. 
Unfortunately, $Q$-learning can take a long time to converge, the reason being that the number of parameters $D_SD_A$ can be very large and that, when a pair $(s,a)$ is observed at time $t$, only one of the entries of the matrix $\hbQ^t \in \reals ^{D_S \times D_A}$ is updated. 

\vspace{.2cm}  
\noindent \textbf{Low-rank via matrix factorization.} 
%One approach to  is to either parametrize or regularize the SA-VF. 
Our contribution to facilitate and accelerate the learning of the SA-VF is to limit its degrees of freedom by promoting low-rank solutions. More specifically, we modify the optimization (estimation) of $\bbQ$ to either force or promote low-rank estimates. Since adding the constraint $\text{rank}(\bbQ)\leq M$ to the optimization in \eqref{E:Q_estimation_plain} --or, alternatively, adding $\text{rank}(\bbQ)$ as a regularizer-- is computationally intractable, one can replace $\text{rank}(\bbQ)$ with the nuclear norm $\|\bbQ\|_*$ , its convex surrogate, or implement a non-convex matrix factorization approach \cite{markovsky2012low}. The latter entails considering the matrices $\bbL \in \reals^{D_S \times M} $ and $\bbR  \in \reals^{M \times D_A}$, write the VF as $\bbQ=\bbL \bbR$, and solving
\begin{eqnarray}\label{E:Q_factor_estimation_plain}
\hspace{-.35cm}&\{ \hbL,\hbR \} \!= \! \argmin_{\bbL,\bbR} \frac{1}{2}\sum_{(s,a)\in \ccalM} (q_s^a - [\bbL\bbR]_{s,a})^2.\;\;&
\end{eqnarray}
The approximated SA-VF is then obtained as $\hbQ=\hbL\hbR$, which is guaranteed to have a rank no larger than $M$. While the factorized problem in \eqref{E:Q_factor_estimation_plain} is non-convex, computationally efficient alternating-minimization schemes that, in the context of matrix completion, are guaranteed to converge to a local minimum can be implemented \cite{markovsky2012low,udell2016generalized}. While different stochastic algorithms can be developed to handle \eqref{E:Q_factor_estimation_plain}, since the cost is quadratic and the problem is bilinear in $\bbL$ and $\bbR$, a stochastic alternating least-squares approach is well motivated. Let $\hbL^t$ and $\hbR^t$ denote the stochastic estimates of $\bbL$ and $\bbR$ at the end of the slot $t$; define the matrix $\bbQ^t \in \reals^{D_S\times D_A}$ as $[\bbQ^t]_{s,a} = [\hbL^{t-1}(\hbR^{t-1})]_{s,a}$ for all $ (s,a) \neq (s_t,a_t)$ and $[\bbQ^t]_{s_t,a_t} = q_s^a = r_t + \gamma \max [\hbL^{t-1}(\hbR^{t-1})]_{s',a}$; and let $k=1,...,K$ be an iteration index. With this notation at hand, 
initialize $\hbR_{[0]}  = \hbR^{t-1}$, set $\bar{\bbQ}=\bbQ^t$, and iterate as

%
%\begin{align}
%[\hbq^{t}_{\ccalS}]=&[[\hbQ^{t-1}]_{1,a_t},..., [\hbQ^{t-1}]_{s_t-1,a_t}, q_{s_t}^{a_t}, [\hbQ^{t-1}]_{s_t+1,a_t}, ..., [\hbQ^{t-1}]_{D_S,a_t}]^T\\ 
%[\hbq^{t}_{\ccalA}]=&[[\hbQ^{t-1}]_{s_t,1},..., [\hbQ^{t-1}]_{s_t,a_t-1}, q_{s_t}^{a_t}, [\hbQ^{t-1}]_{s_t,a_t+1}, ..., [\hbQ^{t-1}]_{s_t,D_A}]^T
%\end{align}
%initialize $[\hbR_{[0]}^t]_{s_t} = [\hbR^{t-1}]_{s_t}$, iterate for $k=1,...,K$ as
%\begin{align} \label{E:LR_learning_factor_alt_LS}
%      [\hbL_{[k]}^t]_{s_t}  =&  (\hbR_{[k-1]}^{t})^{\dagger}\hbq^{t}_{\ccalA} \\
%      [\hbR_{[k]}^t]_{a_t} =&  (\hbL_{[k]}^{t})^{\dagger} \hbq^{t}_{\ccalS}, 
%\end{align}
%and, finally, set the stochastic estimates as $[\hbL^t]_{s_t} = [\hbL_{[K]}^t]_{s_t}$ and $[\hbR^t]_{a_t} = [\hbR_{[K]}^t]_{a_t}$.
%
\begin{align}\label{E:LR_learning_factor_alt_LS}
%\hbL_{[k]} = \bbQ^t(\hbR_{[k-1]}^T)^{\dagger}\;\;\text{and}\;\;\hbR_{[k]} = (\hbL_{[k-1]})^{\dagger} \bbQ^t
\hbL_{[k]} &= \bar{\bbQ}\hbR_{[k-1]}^\Tr(\hbR_{[k-1]}\hbR_{[k-1]}^\Tr)^{-1}\;\;\text{and}\;\;\\
\hbR_{[k]} &=  (\hbL_{[k]}^\Tr\hbL_{[k]})^{-1}\hbL_{[k]}^\Tr\bar{\bbQ}\label{E:LR_learning_factor_alt_LS_line2}
%\;\;\;\text{for}\;\ k=1,...,K; 
\end{align}
for $k=1,...,K$ and, finally, set the stochastic estimates to $\hbL^t = \hbL_{[K]}$ and $\hbR^t = \hbR_{[K]}$. A simpler alternative to optimize \eqref{E:Q_factor_estimation_plain} is to run just a single iteration of a stochastic gradient descent. To that end, let $[\hbL^t]_{s_t}\in \reals^M$ be a vector collecting the $M$ entries of $\hbL^t$ associated with state $s_t$ and, likewise, let $[\hbR^t]_{a_t}\in \reals^M$ be a vector collecting the $M$ entries of $\hbR^t$ associated with action $a_t$. With these  conventions, at time $t$, we use $(s_t,a_t,{s_{t+1}})$ to update the stochastic estimates as follows 
\begin{align}
[\hbL^t]_{s_t} =&  [\hbL^{t-1}]_{s_t} + \alpha_t \big( r_t + \gamma \max_a [\hbL^{t-1}\hbR^{t-1}]_{s_{t+1},a}   \nonumber\\
-& [\hbL^{t-1}\hbR^{t-1}]_{s_t,a_t}\big)[\hbR^t]_{a_t} \label{E:LR_learning_factor_alt_SGD_a}\\
[\hbR^t]_{a_t} =&  [\hbR^{t-1}]_{a_t} + \alpha_t \big( r_t + \gamma \max_a [\hbL^{t-1}\hbR^{t-1}]_{s_{t+1},a}  \nonumber\\ &- [\hbL^{t-1}\hbR^{t-1}]_{s_t,a_t}\big)[\hbL^t]_{s_t}, \label{E:LR_learning_factor_alt_SGD_b}
\end{align}
and $[\hbL^t]_{s}=[\hbL^{t-1}]_{s}$ and $[\hbR^t]_{a}=[\hbR^{t-1}]_{a}$ for all $(s,a)\neq (s_t,a_t)$. The updates in \eqref{E:LR_learning_factor_alt_SGD_a}-\eqref{E:LR_learning_factor_alt_SGD_b} resemble those of the $Q$-learning algorithm in \eqref{E:Q_learning_estimation}, are less computationally demanding than those in \eqref{E:LR_learning_factor_alt_LS}, and for sufficiently small $\alpha_t$ converge to a local optimum. In contrast, the convergence of the alternating scheme (which is not always guaranteed) happens at a faster pace. If the speed of convergence is an issue, the stepsize in \eqref{E:LR_learning_factor_alt_SGD_a}-\eqref{E:LR_learning_factor_alt_SGD_b} can incorporate a gradient normalization to circumvent the slower convergence due to plateaus and saddle points oftentimes present in non-convex optimization \cite{murray2019revisiting}. Finally, all the methods proposed here consider an exploration module to guarantee that all $(s,a)$ pairs are visited. 

Compared to $Q$-learning, the algorithms just described exhibit two main advantages: a) the number of parameters to estimate is smaller --$(D_S + D_A)M$ vs. $D_SD_A$-- and b) at each $t$, the entire row $[\hbL^t]_{s_t}$ and column $[\hbR^t]_{a_t}$ (with $M$ values each) are updated. This contrasts with $Q$-learning, where only a single parameter, the entry $[\hbQ^t]_{s_t,a_t}$, is updated at time $t$.  

%Note also that: 1) as already mentioned, at every $t$ the factorized formulations update multiple entries of the SA-VF and 2) some of the proposed algorithms are non-convex and, as a result, their stochastic trajectories are less predictable. These features in turn imply that the effect (and need for) exploration in this type of algorithms is different from that in $Q$-learning. 
%While certainly of interest, the analysis of these differences is beyond the scope of this conference paper and left as future work.  

\vspace{.2cm}  
\noindent \textbf{Matrix factorization meets the nuclear norm.} Leveraging the fact that the nuclear norm can be written as $\|\bbQ\|_*=\frac{1}{2}\min_{\bbL\bbR=\bbQ} \|\bbL\|_F^2 + \|\bbR\|_F^2$, the cost in \eqref{E:Q_factor_estimation_plain} can be augmented with the Frobenius regularizers $\eta\|\bbL\|_F^2$ and $\eta \|\bbR\|_F^2$. The regularizers are quadratic; hence, the structure and computational complexity of the updated stochastic iterates is roughly the same. On the other hand, since the nuclear norm is convex, the regularizers help to stabilize the behavior of our stochastic algorithms. In fact, conditions under which the Frobenius-regularized iterates converge to the same solution than the nuclear-norm minimization can be rigorously obtained \cite{burer2005local,mardani2013decentralized}. Lastly, the regularized formulation promotes solutions whose rank is typically smaller than $M$. This is important because it bypasses the problem of selecting the value of $M$.

\vspace{.1cm}
\noindent \textbf{Comparison with parametric linear estimation of the SA-VF.} Linear parametric schemes for SA-VF estimation consider the approximation $[\hbQ]_{s,a}\!=\!Q_{\theta}(s, a)\!=\!\bbphi(s, a)^T \bbtheta$, where $\bbphi(s, a)\in \reals^M$ is the set of $M$ features that defines a particular state-action pair and $\bbtheta\in \reals^M$ are the parameters/weights to be estimated online \cite{behzadian2019fast}. The key issue in these approaches is how to pre-design the feature vectors $\bbphi(s, a)$. Differently, we learn the feature vectors and the weights jointly. To be specific, suppose that $\text{rank}(\hbQ)\!=\!M$ and let $\bbv_m$ ($\bbu_m$) be the $m$th left (right) singular vector and $\sigma_m$ the $m$th singular value. Then, we have that $[\hbQ]_{s,a}=\sum_{m=1}^{M} \sigma_m [\bbv_m]_s [\bbu_m]_a$, illustrating that in our approach the features are implicitly defined as $\bbphi(s, a) = [[\bbv_1]_s [\bbu_1]_a,..., [\bbv_M]_s [\bbu_M]_a]^T$ and the linear weights as $\bbtheta = [\sigma_1,...,\sigma_M]^T$, with the difference being that here both $\bbphi(s, a)$ and $\bbtheta$ are learned online.   

\vspace{.1cm}
\noindent \textbf{Reshaping the Q-matrix.} In many applications, the states and, oftentimes, the actions involve multiple variables. To be mathematically precise, suppose that we have $d_S$ state variables (dimensions) and $d_A$ action variables. If the $i$th state variable takes values from the set $\ccalS_i$, the state space is defined as $\ccalS=\ccalS_1 \times ... \times\ccalS_{d_S}$. Analogously, we have that $\ccalA=\ccalA_1 \times ... \times\ccalA_{d_A}$. Even in those cases, the RL literature refers to $Q(s,a)$ as a two-dimensional function, arranging its values in the form of a matrix $\bbQ$ whose rows represent then tuples of states and its columns tuples of actions. Throughout this paper, we kept such a convention and developed algorithms to learn a low-rank approximation of $\bbQ\in\reals^{|\ccalS|\times|\ccalA |}$ , which, because since $|\ccalA|$ is typically much smaller than $|\ccalS|$, is a (sometimes extremely) tall matrix. 

We propose here an alternative low-rank approximation scheme that first  reshapes $\bbQ\in\reals^{|\ccalS|\times|\ccalA |}$ as $\bar{\bbQ}\in\reals^{N_R\times N_C}$, with  $N_RN_C=|\ccalS| |\ccalA |$, and then implements a low-rank decomposition on the reshaped matrix $\bar{\bbQ}$. The idea in the first step is to have $N_R\approx N_C \approx \sqrt{|\ccalS| |\ccalA |}$, so that the reshaped matrix $\bar{\bbQ}$ is approximately square, reducing the number of parameters to be estimated in the low-rank approximation run in the second step from $P_\bbQ=M(|\ccalS|+|\ccalA|)$ to $P_{\bar{\bbQ}}=M(N_C+N_R)\approx 2M\sqrt{|\ccalS| |\ccalA |}$. 
%To be more clear about this latter point, let us assume for simplicity that $\sqrt{|\ccalS| |\ccalA |}$  is a whole number, use $\ccalQ_{|\ccalS| |\ccalA |}$  to denote the set of all matrices with $|\ccalS| |\ccalA |$  entries (including $\bbQ\in\reals^{|\ccalS|\times|\ccalA |}$) and $\tbQ\in\reals^{N_R\times N_C}$  to denote one particular element of $\ccalQ_{|\ccalS| |\ccalA |}$. Then,  it holds that: i) the rank-$K$ approximation of $\tbQ$ requires estimating $P=K(N_C+N_M)$ parameters; and ii) the element of $\ccalQ_{|\ccalS| |\ccalA |}$  for which $P$ is minimized is a square matrix whose number of rows (columns) is $\sqrt{|\ccalS| |\ccalA |}$ . 
For the practical setups where $|\ccalS| \gg |\ccalA |$ we can split the $d_S$ state variables into two sets and consider a $\bar{\bbQ}$ matrix whose rows correspond to an element of $\ccalS_1\times ... \times \ccalS_{d_{S'}}$, whose columns correspond to an element of $\ccalS_{d_{S'}+1}\times...\times\ccalS_{d_{S}}\times \ccalA_1\times ... \times \ccalA_{d_{A}}$, and whose number of columns and rows is approximately the same. Unless prior knowledge exists, the ordering of the state and action variables is arbitrary and, hence,  the assignment of state variable to columns and rows is arbitrary as well, potentially leading to different approximation performances. 

While conceptually simple, the numerical results will demonstrate the benefits and robustness of the proposed approach in standard RL test-cases. Arranging the values of the $Q$-matrix in the form of a tensor with $d_S+d_A$ dimensions and postulating low-rank tensor decomposition algorithms emerge as a natural follow-up research direction.

\section{Numerical experiments}
\vspace{-.1cm}

\begin{figure*}
    \centering
    \includegraphics[width=0.92\linewidth]{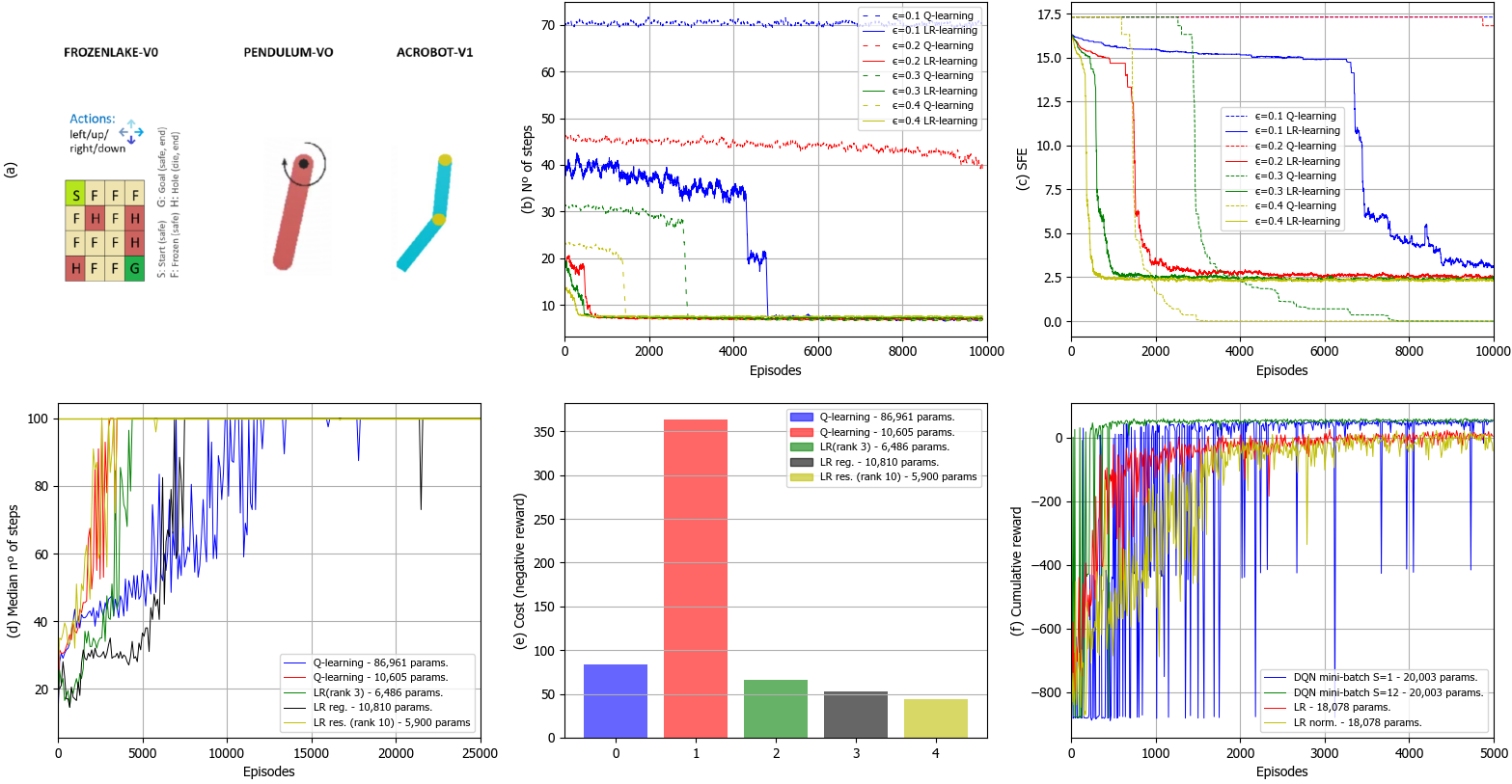}
    \vspace{-.1cm}
    \caption{The upper left picture (a) exemplifies the three tested RL environments, from left to right: FrozenLake-v0, Pendulum-v0 and Acrobot-v1. The two upper right pictures (b, c) show the performance of $Q$-learning and our low-rank (LR) alternative in FrozenLake-v0 for different values of $\epsilon$. The two bottom left pictures (d, e) show the performance of $Q$-learning and the LR alternative in Pendulum-v0. The bottom right picture (f) depicts the performance of DQN and the LR alternative in Acrobot-v1.}
	\vspace{-.6cm}
	\label{fig:results}
\end{figure*}

This section tests the proposed algorithms in three of the standard RL problems of the toolkit OpenAI Gym (see \cite{brockman2016openai} and Fig. \ref{fig:results}.a). Our goal is to illustrate i) the convergence properties of our schemes; ii) the advantages of considering a low-rank $Q(s, a)$ matrix; and iii) the success of our method in an environment too large for classical non-parametric methods. The code and full details can be found in \cite{coderepository}.

\vspace{.1cm}
\noindent \textbf{Convergence properties.} We test the rate of convergence of \eqref{E:Q_factor_estimation_plain} in the FrozenLake-v0 environment \cite{brockman2016openai}. This is a simple deterministic finite-state finite-action grid-like environment, where a reward is given to the agent for reaching a goal state. Although it can be solved using $Q$-learning, our algorithm can exploit the low-rank structure of the environment to accelerate the rate of convergence using almost half of the parameters. 
%
%We compare the rate of convergence of \eqref{E:Q_factor_estimation_plain} to that of $Q$-learning for different exploration probabilities. 
Convergence is analyzed via i) the number of episodes required to obtain the optimal solution;  and ii) measuring the evolution of Squared Frobenius Error ($\mathrm{SFE}$) between the estimated $\hbQ$ and the true $\bbQ$ as $\mathrm{SFE}=||\hbQ - \bbQ||_F^2$. 
%We have considered the true value of $\bbQ$ as the one obtained by the $Q$-learning algorithm after reaching convergence.

We run 100 simulations for different values of $\epsilon_t=\epsilon$. $Q$-learning stores a $16 \times 4$ state-action table, so that the total number of parameters is 64. When testing our low-rank alternative we impose a maximum rank of $M=2$, so that the total number of parameters is $(16+4)M=40$. The two main observations that can be obtained from Fig. \ref{fig:results}.b and \ref{fig:results}.c are that the low-rank alternatives: 1) can converge to the optimal solution, both in terms of the number of steps and $\mathrm{SFE}$; and 2) converge faster in all possible scenarios, that is, for a fixed value of $\epsilon$, the low-rank approach proposed in \eqref{E:Q_factor_estimation_plain} requires fewer episodes than $Q$-learning to converge.

\vspace{.1cm}
\noindent \textbf{Parametrization efficiency.} In the Pendulum-v0 environment \cite{brockman2016openai}, an agent tries to maintain a pendulum upright. This problem has two continuous states (angle and angular velocity), one continuous action, and the reward is multi-objective (keeping the pendulum vertical while minimizing the action exerted). Tackling the continuous-defined problem requires discretizing the state-action space. However, the finer the discretization, the larger the size of the $Q(s, a)$-matrix. 
%Imposing low-rank can drastically reduce the number of parameters while keeping a fine sampling resolution. 
In particular, the Cartesian product of the regularly-sampled states defines a discrete state space of size $D_S=2121$ (see \cite{coderepository} for details). Five RL schemes have been compared with different discretizations of the action space. Two of them are $Q$-learning with the action space discretized at low and high resolutions, with size $D_A=5$ and $D_A=41$ respectively. Two are low-rank schemes that use high-resolution discretization ($D_A=41$ actions), keep the shape of the $Q$ matrix, and consider a (low) rank of $M=3$ and $M=5$, respectively. The last scheme uses the same resolution, but reshapes the $Q$ matrix (with $295$ rows and $295$ columns) and sets the rank to $M=10$. Note that imposing low rank brings down the number of parameters significantly. A roughly-discretized version of $Q$-learning uses $10,605$ parameters while the finer version needs $86,961$ parameters. The low-rank alternative with rank $M=3$ only needs $6,486$ parameters while the larger version with $M=5$ needs $10,810$ parameters. Finally, the reshaped alternative, which is closer to a square matrix, requires $5,900$ parameters for $M=10$. Variants of $Q$-learning undersampling the state space did not converge. Notice also that the roughly-discretized version of the $Q$-learning with $D_A=5$ actions and our low-rank scheme with $M=5$ entail basically the same number of parameters. The low-rank scheme with rank $M=3$ and the reshaped one with $M=10$ were trained with the stochastic algorithm defined in \eqref{E:LR_learning_factor_alt_SGD_a} and \eqref{E:LR_learning_factor_alt_SGD_b} while the low-rank scheme with the larger rank $M=5$ was trained with the regularized alternative. A certain level of exploration is required, thus the exploratory probability was set to $\epsilon_t=0.2$. 

Each algorithm was trained $10$ times, and the results obtained can be found in Figs. \ref{fig:results}.d and \ref{fig:results}.e. These results are obtained running a greedy test episode every several training episodes. As expected, the smaller version of $Q$-learning converges faster to a solution able to keep the pendulum upright, but fails at selecting low-cost actions (the poor discretization of the action space forces the agent to select sub-optimal actions). In contrast, the proposed low-rank schemes converge faster than the $Q$-learning version with the largest number of parameters. For the simulated number of episodes, they also outperform both versions of $Q$-learning in terms of cost. Moreover, an SVD decomposition of the $Q$-matrix estimated by the $Q$-learning algorithm reveals that the first three singular values account for more than $80$ \% of the total variance. On the other hand, the reshaped scheme is the best performing agent. A more prudently designed parametrization of the model leads to better results in terms of both, cumulative rewards and speed of convergence. These results demonstrate that the (unconstrained) optimal solutions are inherently low-rank and, hence, our algorithms fruitfully exploit this structure. 

\noindent \textbf{Comparison against the state-of-the-art.} The Acrobot-v1 is a double-pendulum environment that mimics a gymnast trying to swing up \cite{brockman2016openai}. There are six state variables: the sine and cosine of the angle and the angular velocity of the upper and lower joints. Only one action variable exists, which can take three different values. The number of discrete states produced by the Cartesian product of regularly-sampled states grows exponentially on the number of states. Even discretizing each of the six states at low-resolution results in roughly $7\cdot 10^6$ states, leading to a $Q(s,a)$-matrix with more than $20\cdot 10^6$ entries (see \cite{coderepository} for details). This dimensionality explosion can be addressed using an intelligent reshaping of the $Q(s, a)$ into a $4520 \times 4519$ matrix and applying the LR scheme imposing low rank $M=2$. The total number of parameters drops dramatically to $18,078$. Two versions of the LR algorithm are compared in this experiment, one that normalizes the stochastic gradient update and another one that does not. 

The VF and SA-VF-based approaches to handle high-dimensional environments are limited to parametric estimators. Here we will compare our performance with that of Deep $Q$-learning (DQN)~\cite{mnih2013playing}, an offline NN state-of-the-art representative of parametric value-based methods. Linear methods were also tested but their performance is not shown because their value was worse and required many more episodes (more than one additional order of magnitude) to converge. NN-based RL algorithms suffer from convergence problems due to the temporal correlation. This is typically overcome by using experience replay (ER)~\cite{mnih2013playing}, which requires the training to be offline. A one-layer Multi-Layer Perceptron (MLP) with a comparable number of parameters is used to implement DQN. One simulation implements a more efficient version of DQN using mini-batches of size $S=12$ obtained from the ER buffer to perform each training update. A lighter version of DQN is trained using mini-batches of size $S=1$.

The results of 10 trials are shown in Fig. \ref{fig:results}.f. As expected, the \textit{parametric} NN-based schemes converge to a better solution. However, our normalized \textit{non-parametric} LR alternative successfully handles the problem and, in fact, converges faster than the light version of DQN. In contrast to DQN, the LR algorithm is fully online and does not need ancillary offline twists to stabilize the training. It is also important to note that the DQN trained with mini-batches of size $S=12$ took 2.5 more training time: the use of larger batches introduces a trade-off between the training time and efficient use of previous experience. The effect of the normalization of the gradient can be observed as well. Both LR variants converge to the same solution, but the normalized one does it faster.

\section{Concluding remarks}

A collection of RL algorithms that exploit the intrinsic low-rank structure of the SA-VF has been proposed. The schemes leverage sounded matrix completion results to estimate the $Q$-function in a \textit{model-free} and \textit{online} fashion. Moreover, a reshaping of the $Q$-matrix has been introduced to parametrize the model in a more efficient way. The numerical experiments show the advantages of the low-rank models both in terms of speed of convergence and the achieved cumulative reward. 

%\newpage

%\section{Concluding summary}

%Leveraging results from low-rank minimization and matrix completion, this paper presented several (model-free) RL learning algorithms for the stochastic estimation of the $Q$-function. Two estimation problems were formulated. In the first one, the state-action value matrix is factorized as the product of a tall and a fat matrix, and the focus is on using stochastic optimization to update those factors based on the current state, action and reward, as well as the state at the next time instant. In the second formulation, two Frobenius regularizers were added to the objective, so that the factorized approach can be connected to imposing a nuclear-norm regularizer on the state-action value matrix. Compared to the classical, non-regularized, $Q$-learning algorithm, the novel schemes entail less parameters. Furthermore, for each state/action/reward/state sample, the proposed schemes updated the value all the entries of the $Q$-function involving \emph{either} the current state or the current action. This contrasts with $Q$-learning, where only the entry involving \emph{both} the current state and action is updated. Numerical simulations run on two two classical RL benchmarks (FrozenLake-v0 and Pendulum-v0) confirmed the practical value of the proposed schemes, both in terms of speed of convergence (for the same aggregated reward) as well as of superior reward (for a reduced number of model parameters).

%\section*{References}
%\newpage
\bibliographystyle{IEEEtran}
\bibliography{references}

\begin{thebibliography}{10}
\providecommand{\url}[1]{#1}
\csname url@samestyle\endcsname
\providecommand{\newblock}{\relax}
\providecommand{\bibinfo}[2]{#2}
\providecommand{\BIBentrySTDinterwordspacing}{\spaceskip=0pt\relax}
\providecommand{\BIBentryALTinterwordstretchfactor}{4}
\providecommand{\BIBentryALTinterwordspacing}{\spaceskip=\fontdimen2\font plus
\BIBentryALTinterwordstretchfactor\fontdimen3\font minus
  \fontdimen4\font\relax}
\providecommand{\BIBforeignlanguage}[2]{{%
\expandafter\ifx\csname l@#1\endcsname\relax
\typeout{** WARNING: IEEEtran.bst: No hyphenation pattern has been}%
\typeout{** loaded for the language `#1'. Using the pattern for}%
\typeout{** the default language instead.}%
\else
\language=\csname l@#1\endcsname
\fi
#2}}
\providecommand{\BIBdecl}{\relax}
\BIBdecl

\bibitem{sutton2011reinforcement}
R.~S. Sutton and A.~G. Barto, ``Reinforcement learning: An introduction,''
  2011.

\bibitem{bertsekas2019reinforcement}
D.~P. Bertsekas, \emph{Reinforcement learning and optimal control}.\hskip 1em
  plus 0.5em minus 0.4em\relax Athena Scientific, 2019.

\bibitem{mnih2013playing}
V.~Mnih, K.~Kavukcuoglu, D.~Silver, A.~Graves, I.~Antonoglou, D.~Wierstra, and
  M.~Riedmiller, ``Playing {A}tari with deep reinforcement learning,''
  \emph{arXiv preprint arXiv:1312.5602}, 2013.

\bibitem{eckart1936approximation}
C.~Eckart and G.~Young, ``The approximation of one matrix by another of lower
  rank,'' \emph{Psychometrika}, vol.~1, no.~3, pp. 211--218, 1936.

\bibitem{markovsky2012low}
I.~Markovsky, \emph{Low rank approximation}.\hskip 1em plus 0.5em minus
  0.4em\relax Springer, 2019.

\bibitem{watkins1992q}
C.~J. Watkins and P.~Dayan, ``Q-learning,'' \emph{Machine learning}, vol.~8,
  no. 3-4, pp. 279--292, 1992.

\bibitem{udell2016generalized}
M.~Udell, C.~Horn, R.~Zadeh, S.~Boyd \emph{et~al.}, ``Generalized low rank
  models,'' \emph{Foundations and Trends{\textregistered} in Machine Learning},
  vol.~9, no.~1, pp. 1--118, 2016.

\bibitem{mardani2013decentralized}
M.~Mardani, G.~Mateos, and G.~B. Giannakis, ``Decentralized
  sparsity-regularized rank minimization: Algorithms and applications,''
  \emph{IEEE Trans. Signal Processing}, vol.~61, no.~21, pp. 5374--5388, 2013.

\bibitem{barreto2016incremental}
A.~M. Barreto, R.~L. Beirigo, J.~Pineau, and D.~Precup, ``Incremental
  stochastic factorization for online reinforcement learning,'' in \emph{Proc.
  AAAI Conf. Artificial Intelligence}, 2016.

\bibitem{jiang2017contextual}
N.~Jiang, A.~Krishnamurthy, A.~Agarwal, J.~Langford, and R.~E. Schapire,
  ``Contextual decision processes with low bellman rank are pac-learnable,'' in
  \emph{Proc. Intl. Conf. Machine Learning-Volume 70}.\hskip 1em plus 0.5em
  minus 0.4em\relax JMLR. org, 2017, pp. 1704--1713.

\bibitem{ong2015value}
H.~Y. Ong, ``Value function approximation via low-rank models,'' \emph{arXiv
  preprint arXiv:1509.00061}, 2015.

\bibitem{cheng2016co}
B.~Cheng and W.~B. Powell, ``Co-optimizing battery storage for the frequency
  regulation and energy arbitrage using multi-scale dynamic programming,''
  \emph{IEEE Trans. Smart Grid}, vol. 9.3, pp. 1997--2005, 2016.

\bibitem{cheng2017low}
B.~Cheng, T.~Asamov, and W.~B. Powell, ``Low-rank value function approximation
  for co-optimization of battery storage,'' \emph{IEEE Trans. Smart Grid}, vol.
  9.6, pp. 6590--6598, 2017.

\bibitem{yang2019harnessing}
Y.~Yang, G.~Zhang, Z.~Xu, and D.~Katabi, ``Harnessing structures for
  value-based planning and reinforcement learning,'' \emph{arXiv preprint
  arXiv:1909.12255}, 2019.

\bibitem{shah2020sample}
D.~Shah, D.~Song, Z.~Xu, and Y.~Yang, ``Sample efficient reinforcement learning
  via low-rank matrix estimation,'' \emph{arXiv preprint arXiv:2006.06135},
  2020.

\bibitem{chen2018factorized}
Y.~Chen, M.~Zhou, Y.~Wen, Y.~Yang, Y.~Su, W.~Zhang, D.~Zhang, J.~Wang, and
  H.~Liu, ``Factorized {Q}-learning for large-scale multi-agent systems,''
  \emph{arXiv preprint arXiv:1809.03738}, 2018.

\bibitem{melo2007q}
F.~S. Melo and M.~I. Ribeiro, ``Q-learning with linear function
  approximation,'' in \emph{Intl. Conf. Comp. Learning Theory}.\hskip 1em plus
  0.5em minus 0.4em\relax Springer, 2007, pp. 308--322.

\bibitem{behzadian2019fast}
B.~Behzadian, S.~Gharatappeh, and M.~Petrik, ``Fast feature selection for
  linear value function approximation,'' in \emph{Proc. Intl. Conf. Automated
  Planning and Scheduling}, vol.~29, no.~1, 2019, pp. 601--609.

\bibitem{behzadian2018low}
B.~Behzadian and M.~Petrik, ``Low-rank feature selection for reinforcement
  learning.'' in \emph{ISAIM}, 2018.

\bibitem{murray2019revisiting}
R.~{Murray}, B.~{Swenson}, and S.~{Kar}, ``Revisiting normalized gradient
  descent: Fast evasion of saddle points,'' \emph{IEEE Trans. Automatic
  Control}, vol.~64, no.~11, pp. 4818--4824, 2019.

\bibitem{burer2005local}
S.~Burer and R.~D. Monteiro, ``Local minima and convergence in low-rank
  semidefinite programming,'' \emph{Mathematical Programming}, vol. 103, no.~3,
  pp. 427--444, 2005.

\bibitem{brockman2016openai}
G.~Brockman, V.~Cheung, L.~Pettersson, J.~Schneider, J.~Schulman, J.~Tang, and
  W.~Zaremba, ``Open{AI} {G}ym,'' \emph{arXiv preprint arXiv:1606.01540}, 2016.

\bibitem{coderepository}
V.~T. S.~Rozada and A.~G. Marques, ``Online code repository: Low-rank
  state-action value-function approximation,''
  \emph{https://github.com/sergiorozada12/low-rank-rl}.

\end{thebibliography}

\end{document}